\documentclass[letterpaper, 10 pt, conference]{ieeeconf}  

\IEEEoverridecommandlockouts                              
\usepackage{graphics} 
\usepackage{times} 
\usepackage{amsmath,amsfonts} 
\usepackage{amssymb}  
\usepackage[caption=false,font=scriptsize,labelfont=sf,textfont=sf]{subfig}
\usepackage{textcomp}
\usepackage{stfloats}
\usepackage{url}
\usepackage{verbatim}
\usepackage{graphicx}
\usepackage{booktabs}
\usepackage{multirow}
\usepackage{makecell}
\usepackage{hyperref}
\usepackage{cite}
\usepackage{siunitx}
\usepackage{booktabs}
\usepackage{makecell}
\DeclareSubrefFormat{myparens}{#1(#2)}
\bstctlcite{IEEEexample:BSTcontrol}
\usepackage{algorithmic}
\usepackage[ruled]{algorithm2e}
\usepackage[table]{xcolor}
\usepackage{hyperref}
\hypersetup{
    colorlinks = true,   
    urlcolor   = blue,   
    linkcolor  = red,    
    citecolor  = green   
}

\definecolor{codegreen}{rgb}{0.0,0.6,0.0}

\title{\LARGE \bf USVTrack: USV-Based 4D Radar-Camera Tracking Dataset\\ for Autonomous Driving in Inland Waterways}
\author{
Shanliang Yao$^{1}$, Runwei Guan$^{2}$, Yi Ni$^{3}$, Sen Xu$^{4}$, Yong Yue$^{3}$, Xiaohui Zhu$^{3, \dagger}$, Ryan Wen Liu$^{5, \dagger}$
\thanks{This work was supported in part by Suzhou Municipal Key Laboratory for Intelligent Virtual Engineering (NO. SZS2022004), XJTLU AI University Research Centre, Jiangsu Province Engineering Research Centre of Data Science and Cognitive Computation at XJTLU and SIP AI innovation platform (NO. YZCXPT2022103), National Natural Science Foundation of China (NO. 52422111) and Fund of Hubei Key Laboratory of Inland Shipping Technology (NO. NHHY2024004).}
\thanks{$^{\text{1}}$ Shanliang Yao is with the School of Information Engineering, Yancheng Institute Technology, Yancheng 224051, China, and also with Hubei Key Laboratory of Inland Shipping Technology, Wuhan 430063, China. (email: shanliang.yao@ycit.edu.cn).}
\thanks{$^{\text{2}}$ Runwei Guan is with Thrust of Artificial Intelligence and Thrust of Intelligent Transportation, The Hong Kong University of Science and Technology (Guangzhou), Guangzhou 511400, China. (email: runweiguan@hkust-gz.edu.cn).}
\thanks{$^{3}$ Yi Ni, Yong Yue and Xiaohui Zhu are with School of Advanced Technology, Xi'an Jiaotong-Liverpool University, Suzhou 215123, China. (email: yi.ni21@student.xjtlu.edu.cn, \{yong.yue, xiaohui.zhu\}@xjtlu.edu.cn).}
\thanks{$^{\text{4}}$ Sen Xu is with the School of Information Engineering, Yancheng Institute Technology, Yancheng 224051, China. (email: xusen@ycit.cn).}
\thanks{$^{\text{5}}$ Ryan Wen Liu is with School of Navigation, Wuhan University of Technology, Wuhan 430063, China, and also with the State Key Laboratory of Maritime Technology and Safety, Wuhan 430063, China. (email: wenliu@whut.edu.cn).}
\thanks{$^{\dagger}$ Corresponding author: xiaohui.zhu@xjtlu.edu.cn, wenliu@whut.edu.cn}
}

\begin{document}

\maketitle
\thispagestyle{empty}
\pagestyle{empty}

\begin{abstract}
Object tracking in inland waterways plays a crucial role in safe and cost-effective applications, including waterborne transportation, sightseeing tours, environmental monitoring and surface rescue. Our Unmanned Surface Vehicle (USV), equipped with a 4D radar, a monocular camera, a GPS, and an IMU, delivers robust tracking capabilities in complex waterborne environments. By leveraging these sensors, our USV collected comprehensive object tracking data, which we present as USVTrack, the first 4D radar-camera tracking dataset tailored for autonomous driving in new generation waterborne transportation systems. Our USVTrack dataset presents rich scenarios, featuring diverse various waterways, varying times of day, and multiple weather and lighting conditions. Moreover, we present a simple but effective radar-camera matching method, termed RCM, which can be plugged into popular two-stage association trackers. Experimental results utilizing RCM demonstrate the effectiveness of the radar-camera matching in improving object tracking accuracy and reliability for autonomous driving in waterborne environments. The USVTrack dataset is public on \url{https://usvtrack.github.io}.
\end{abstract}

\section{Introduction}

In recent years, rapid advancement of autonomous driving technologies has garnered significant attention. While substantial progress has been made in road-based autonomous systems, the importance of autonomous driving employing Unmanned Surface Vehicles (USVs) in inland waterways cannot be overstated \cite{yan2021architecture, bovcon2019mastr1325}. As a core component of these systems, USVs leverage all-electric drivetrain and autonomous driving technology to deliver economic, safe, and efficient transportation solutions. USVs are widely utilized in a variety of sustainable applications, ranging from cargo transport and sightseeing tours to waste cleaning, water rescue, and environmental monitoring \cite{yao2024waterscenes, cheng2021we, kiefer20253rd}.

\begin{figure}[!t]
\begin{center}
\includegraphics[width=1\linewidth]{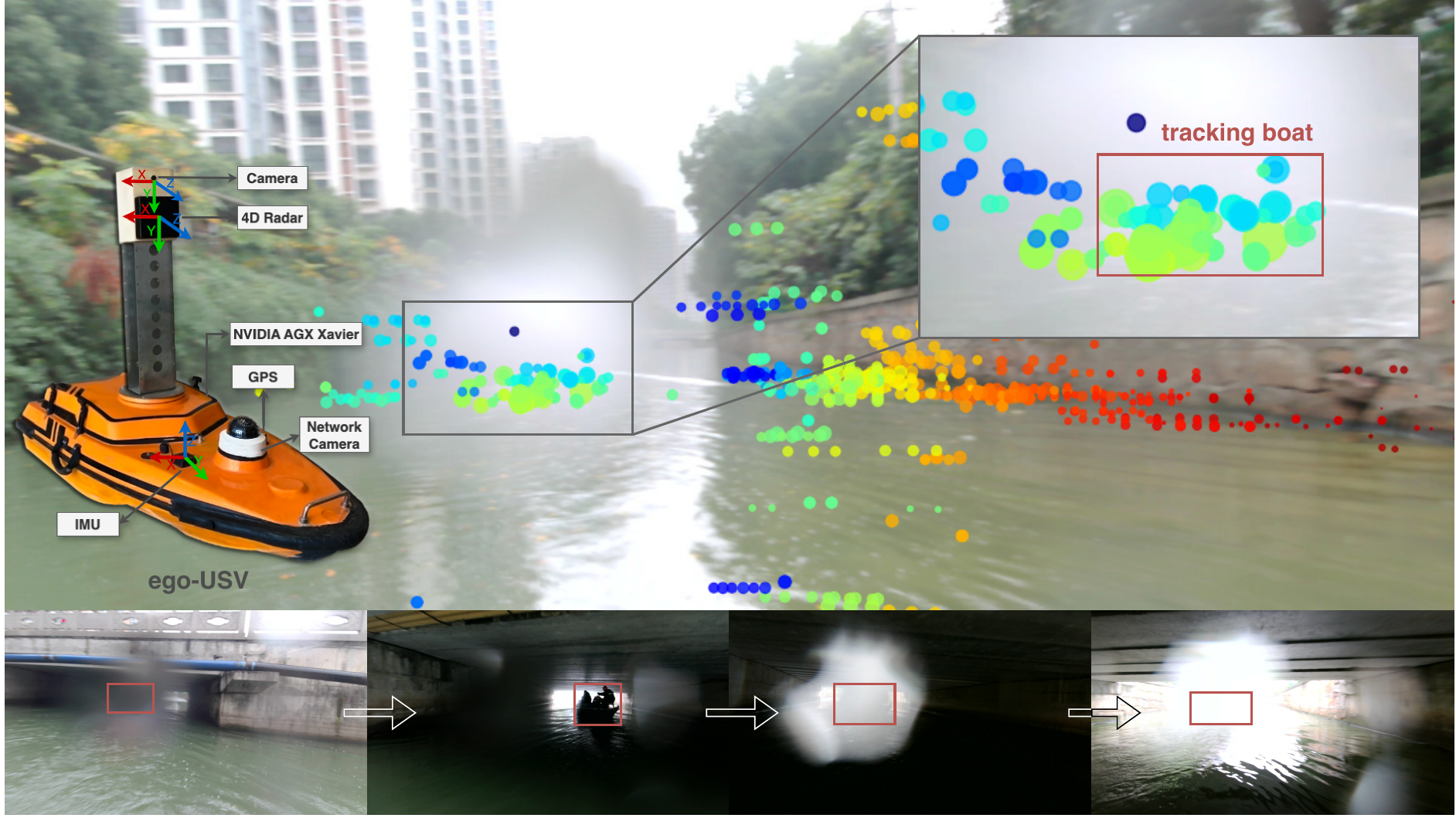}
\end{center}
\vspace{-4mm}
\caption{Example scenario from our USVTrack dataset from the viewpoint of our ego-USV. For each radar point on the image, the color indicates the range, while the size represents reflected power from the target. (a) The radar sensor detects the boat in scenarios where the camera is occluded by water droplets. (b) The four images below depict the tracked boat navigating through a bridge hole, with the camera sensor being impacted in such situation.}
\label{fig:cover}
\end{figure}

\begin{table*}[!h]
\caption{Overview of public tracking datasets in waterborne environments.}
\center
\footnotesize
\begin{tabular*}{\linewidth}{p{2.2cm}<{}p{0.4cm}<{\centering}p{2.2cm}<{\centering}p{1.85cm}<{\centering}p{0.8cm}<{\centering}p{2.9cm}<{\centering}p{2.18cm}<{\centering}p{2.2cm}<{\centering}}
\toprule

\bf{Name} & \bf{Year} & \bf{Sensors} & \bf{Platform} & \bf{Classes} & \bf{Videos, Frames} & \bf{Waterways} & \bf{Conditions}\\\midrule
SMD \cite{moosbauer2019benchmark} & 2019 & Camera & On-shore/vessel & 10 & 63 clips, 31,653 frames & Maritime & Lighting \\
FloW \cite{cheng2021flow} & 2021 & Camera, 3D Radar & USV & 1 & 200 clips, no annotations & River & Lighting \\
SeaDronesSee \cite{varga2022seadronessee} & 2022 & Camera  & UAV & 2 & 22 clips, 54,105 frames & Maritime & Lighting \\
BoaTrack \cite{kiefer20242nd} & 2023 & Camera & USV & 1 & 3 clips, no annotations & Maritime & - \\
\midrule
\textbf{USVTrack} (Ours) & 2025 & Camera, 4D Radar, GPS, IMU & USV & 3 & 60 clips, 68,822 frames, 85,229 annotations & River, Lake, Canal, Moat, Dock & Lighting, Weather, Seasonality  \\
\bottomrule
\end{tabular*}
\label{tab:related-datasets}
\end{table*}

Despite the broad potential for USV-based waterborne transportation systems, inland waterways present a complex and challenging environment. These waterways are characterized by narrow river, varying water levels, and dynamic obstacles such as floating debris, bridges, and other vessels.
Camera-based perception for USVs still faces numerous challenges in these environments, particularly in the context of object tracking tasks. 
As is demonstrated in Fig. \ref{fig:cover}, water droplets and vapor accumulated on the camera lens can obstruct the view of surrounding objects, resulting in blurred or obscured view of targets.
Factors such as the wind and waves in waterborne environments can cause both the USV and the target to sway, making unstable object tracking when captured by the camera.
All these manifold factors present a series of challenges to the camera sensor, complicating the detection and tracking of objects in aquatic surroundings.

To address these challenges, there has been a growing interest in integration additional perception sensors into autonomous driving systems. Radar sensors, which emit radio waves that bounce off objects and return to the sensor, offer distinct advantages in measuring range, Doppler velocity, and reflected power \cite{yao2025exploring}. The longer wavelength of millimeter waves allows them to penetrate through atmospheric phenomena, ensuring reliable and consistent performance in challenging weather scenarios \cite{bijelic2020seeing}. Unlike cameras that can be affected by variations in lighting conditions, radar sensors operate independently of ambient light, maintaining consistent performance under strong sunlight or low visibility \cite{bilik2022comparative}. All these advantages make radar sensors a reliable and robust component in autonomous driving vehicles, equally well-suited to overcoming the unique challenges of waterborne environments.

In recent years, radar sensors have revolutionized from 3D (range, Doppler velocity, azimuth angle) to 4D, with the additional elevation angle. This advancement enables 4D radars to achieve higher resolution and generate denser point clouds, providing richer information about targets \cite{liu2023smurf, yang2024v2x, ding2025radarocc}. 
Numbers of radar-camera tracking dataset are proposed for autonomous driving on roads, (e.g., K-Radar \cite{paek2022k}, TJ4DRadSet \cite{zheng2022tj4dradset}, Dual Radar\cite{zhang2023dual}, and NTU4DRadLM \cite{zhang2023ntu4dradlm}), and numerous studies demonstrate that 4D radar-camera fusion can improve the accuracy and robustness of tracking \cite{yao2023radar, bai2024sgdet3d, xiong2025lxlv2}.

However, 4D radar-camera tracking in waterborne transportation systems remains unexplored, primarily due to the lack of available pioneering datasets and methods. 
To address this research gap, our study contributes to the field by introducing a USV-based 4D radar-camera tracking dataset and proposing a radar-camera matching method to achieve higher data association in the challenging  waterway environments. 
Our main contributions are listed as follows:

\begin{itemize}
\item We present USVTrack, the first USV-based 4D radar-camera tracking dataset for autonomous driving in waterborne transportation systems, providing comprehensive data from four sensors: a 4D radar, a monocular optical camera, a GPS and an IMU.
\item USVTrack contains a rich diversity of data samples, including various waterways (wide and narrow rivers, lakes, canals, moats and docks), diverse time conditions (daytime, nightfall, night), weather conditions (sunny, overcast, rainy, snowy), and lighting conditions (normal, dim, strong).
\item We propose a simple yet effective radar-camera matching strategy named RCMatch. Experimental results demonstrate that RCMatch improves matching accuracy, thereby enhancing tracking robustness compared to commonly used camera-based trackers.
\end{itemize}

\section{Related Work}
\label{sec:Related Work}

\subsection{Object Tracking Datasets in Waterborne Environments}

TABLE \ref{tab:related-datasets} provides an overview of public object tracking datasets related to waterborne environments.
SMD \cite{moosbauer2019benchmark} dataset contains videos for various obstacle categories in maritime environments, including ferry, ship, vessel, speed boat, and sail boat, acquired from both on-shore and on-vessel perspectives. Besides, some data are captured from a near-infrared camera, enabling image capture in low-light or dark conditions. To draw attention to floating waste cleaning using USVs, FloW \cite{cheng2021flow} dataset is proposed for floating waste detection task. Using both camera and radar sensors, benchmark evaluations on this dataset highlighted the radar sensor's effectiveness in detecting small objects. However, for the object tracking task, the Flow dataset only offers 200 clips without annotations, and is limited to a single category (``bottle"), restricting its applicability for autonomous driving on inland waterways.

The SeaDronesSee \cite{varga2022seadronessee} dataset is a UAV-based visual dataset specifically designed for detecting and tracking humans in open water.  It includes two distinct sub-tasks for object tracking: MOT-Swimmer, which focuses on tracking persons in water (floaters and swimmers), and MOT-All-Objects-In-Water, which tracks all objects in water (excluding people on boats). However, in both tasks, all objects are grouped into a single class.
Introducing a dataset from the perspective of USVs, BoaTrack \cite{kiefer20242nd} is a USV-based dataset aimed at detecting and tracking boats and other objects (such as buoys) in open water and dock scenarios. However, it does not provide training or validation videos and focuses solely on tracking boats, limiting its scope for broader applications.

\subsection{Object Tracking Methods}

Tracking-By-Detection (TBD) is a popular paradigm in object tracking, where the tracking process revolves around detecting objects in each frame and associating them across frames using matching strategies to maintain consistent tracks. Object detection architectures such as Faster R-CNN \cite{ren2015faster} and YOLO series \cite{wang2023yolov7, yolov8, tian2025yolov12} are employed to identify and locate objects within individual frames of a video sequence. Robust data association methods are then leveraged to link object detections across frames. Beyond traditional methods like the Hungarian algorithm \cite{kuhn1955hungarian}, Kalman filter \cite{welch1995introduction}, deep learning techniques are used to calculate feature similarities for more accurate association, including appearance-based tracking methods (e.g., DeepSORT \cite{wojke2017simple}) and motion-based tracking methods (e.g., ByteTrack \cite{zhang2022bytetrack}, BoT-SORT \cite{aharon2022bot}, OC-SORT \cite{cao2023observation}). 

Joint Detection and Tracking (JDT) is a paradigm that simultaneously accomplishes object detection and identity embedding features within a unified framework. Methods such as FairMOT \cite{zhang2021fairmot}, Tracktor++ \cite{bergmann2019tracking} further explore the joint training of object detection and appearance embedding. Transformer-based methods utilize powerful attention mechanism to model the association task, including TrackFormer \cite{meinhardt2022trackformer}, MOTR series \cite{zeng2022motr, zhang2023motrv2, yu2023motrv3} and MOTIP \cite{gao2024multiple}. However, the Transformer-based matchers involve a significant number of self/cross-attention operations, which increase computational complexity and hinder real-time performance.

\section{USVTrack Dataset} 

\begin{figure*}[htbp]
\begin{center}
\includegraphics[width=1\linewidth]{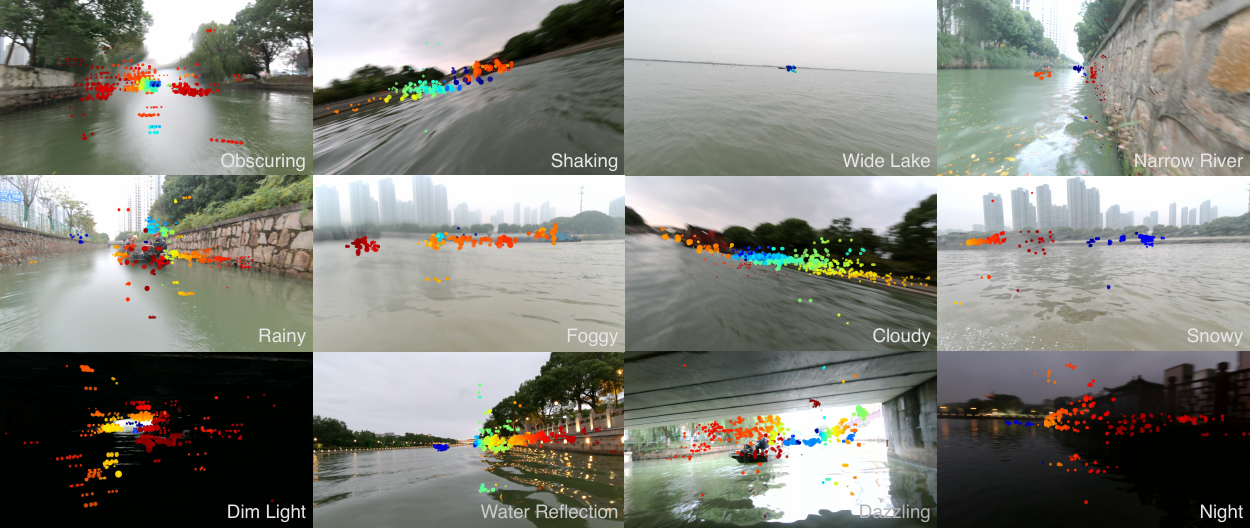}
\end{center}
\caption{Samples in USVTrack dataset. Radar points are projected onto the image plane as colored dots. For each radar point on the image, the color denotes the range, and the size represents reflected power from the target.}
\label{fig:samples}
\end{figure*}

USVTrack dataset is the first 4D radar-camera tracking dataset in waterborne transportation systems, incorporating data from 4D radar, camera, GPS, and IMU sensors.
As can be intuitively seen from Fig. \ref{fig:samples}, our USVTrack dataset includes common types of vessels under various waterway, lighting, weather, and seasonal conditions.
Key details about the USVTrack dataset, such as equipped sensors, acquisition platform, object categories, dataset size, and collection conditions, are summarized in TABLE \ref{tab:related-datasets}. In this section, we outline the process of creating the dataset and provide a comprehensive statistical analysis of the collected data.

\subsection{USV Platform}

Our USV platform, depicted in Fig. \ref{fig:cover}, features the origin and direction of each sensor represented by different colors in the coordinate systems. The USV is equipped with a monocular optical camera for capturing image information, a 4D radar for collecting radar point cloud data, a GPS for geographic location information, and an IMU for tracking the USV's motion and posture. Additionally, a network camera with a 360-degree viewing angle is integrated for panoramic environment monitoring. To support automatic navigation and data collection, an NVIDIA AGX Xavier is deployed in the cabin with customized scripts. Detailed specifications of the sensors are outlined in TABLE \ref{tab:sensors}.

\begin{table}[hbt]
\caption{Specifications of sensors equipped on our USV.}
\footnotesize
\begin{tabular*}{\linewidth}{p{0.8cm}  p{7.2cm}}
\toprule
\bf{Sensor} & \bf{Specifications} \\\midrule
\textbf{Radar} & Type: Oculii EAGLE $77\text{GHz}$ Point Cloud Radar; Mode: Medium Range 200m; Range resolution: 0.43m; Velocity resolution: 0.27m/s; Azimuth/elevation angle resolution: $<$1$^\circ$; HFOV: 110$^\circ$; VFOV: 45$^\circ$; Frequency: 15Hz. \\
\vspace{-1mm} \textbf{Camera} & \vspace{-1mm} Type: SONY IMX317 CMOS sensor; Resolution: 1920 $\times$ 1080; HFOV: 100$^\circ$; VFOV: 60$^\circ$; Frequency: 30Hz.\\
\vspace{-1mm} \textbf{GPS} & \vspace{-1mm} Coordinates: latitude, longitude and altitude; Position accuracy: $<$2.5m; Velocity accuracy: $<$0.1m/s; Frequency: 10Hz.\\
\vspace{-1mm} \textbf{IMU} & \vspace{-1mm} 10-axis: 3-axis gyroscope, 3-axis accelerometer, 3-axis magnetometer and a barometer; Heading accuracy: $0.5^\circ$; Roll/pitch accuracy: $0.1^\circ$; Frequency: 50Hz.\\
\bottomrule
\end{tabular*}
\label{tab:sensors}
\end{table}

\subsection{Data Collection}

The dataset collection process was meticulously executed, encompassing a diverse range of waterways, lighting conditions, weather patterns, and seasonal variations. This comprehensive approach ensures the dataset's diversity and adaptability to real-world scenarios, enhancing its utility for developing robust and generalizable autonomous driving solutions in waterborne environments.

\subsubsection{Waterway Diversity}
Data collection for the USVTrack dataset spanned a variety of waterways, including wide rivers, narrow rivers, lakes, canals, moats, and docks. This geographical diversity is crucial for waterborne transportation systems, as each waterway type presents distinct structures, transportation vehicles, and surrounding environments, ensuring the dataset's applicability to a wide range of scenarios.

\subsubsection{Lighting Conditions}
The USVTrack dataset incorporates data captured under diverse lighting conditions, ranging from bright daylight to low-light scenarios and challenging nighttime conditions. This diversity enables a comparative analysis of radar and camera sensor characteristics, facilitating the development of systems capable of effective navigation across varying natural light conditions.

\subsubsection{Weather Variability}
The data collection process was intentionally conducted under various weather conditions, including clear skies, overcast conditions, and adverse weather events such as rain, fog, and snow. While these conditions reduce visibility for camera sensors, the radar sensor's penetration capability ensures reliable object tracking, highlighting its robustness in challenging environments.

\subsubsection{Seasonal Variations}
The USVTrack dataset includes data collected throughout summer, fall, winter, and spring. Seasonal variations introduce unique challenges, such as water vapor due to temperature fluctuations, which can affect sensor performance. By incorporating data from all seasons, the dataset supports the development of algorithms capable of adapting to environmental changes, ensuring consistent tracking performance year-round.

\subsection{Data Annotations}
In USVTrack dataset, we meticulously annotated three common categories in waterborne transportation: ship, boat and vessel. For each frame in the video sequences, we annotated each category with precise 2D bounding boxes. To support the study of object tracking over time, we assigned a unique tracking ID to each annotated target. 
For point clouds in the radar data, we projected the radar data onto the camera plane and annotated the categories and tracking IDs for the point clouds that fall within the image bounding boxes \cite{yao2024waterscenes}.
Furthermore, each video in USVTrack is annotated with scenario attributes, including waterway, lighting, and weather conditions. These detailed attributes further enhance the contextual richness of the dataset and help to understand the impact of different factors on object tracking.

\subsection{Dataset Statistics}
USVTrack dataset comprises 60 clips of camera videos along with corresponding radar point clouds, GPS, and IMU sensor data. Each frame within the camera video has a resolution of 1920 $\times$ 1080 pixels and is presented at a frequency of 30Hz. Following the methodology of the MOT20 \cite{dendorfer2020mot20} dataset, we allocated half of the videos to the training set and the remaining half to the test set. This division was carefully executed to ensure similarity between the subsets in terms of video duration, object count, scene variety, and motion diversity, thereby maintaining a balanced and representative dataset for training and evaluation purposes.

\begin{table}[!h]
\caption{Dataset statistics for camera and radar modalities in USVTrack dataset. The unit for average power is $dB$, and the unit for average velocity is $m/s$.}
\setlength\tabcolsep{5pt} 
\center
\footnotesize
\begin{tabular*}{\linewidth}{p{1cm}p{2cm}<{}|p{1cm}<{\centering}p{0.95cm}<{\centering}p{0.9cm}<{\centering}p{0.95cm}<{\centering}}
\toprule

\bf{Modality} & \bf{Attribute} & \bf{Overall} & \bf{Ship} &  \bf{Boat} & \bf{Vessel} \\\midrule
\multirow{3}{*}{\bf{Camera}} &
Frames & 68,822 & 25,203 & 26,749 & 17,371  \\
& Objects & 85,229 & 35,684 & 30,821 & 18,724 \\
& IDs & 98 & 41 & 38 & 19\\\midrule

\multirow{6}{*}{\bf{Radar}} 
& Frames & 45,091 & 15,614 & 15,189 & 10,101 \\
& Objects & 44,271 & 17,873 & 15,957 &  10,441\\
& IDs & 89 & 32 & 38 & 19 \\\cmidrule(lr){2-6}
& Average Points & 57.22 & 70.80 & 30.02 & 75.56\\
& Average Power & 13.36 & 14.57 & 11.62 & 13.96 \\
& Average Velocity & 1.32 & 0.97 & 0.90 & 2.57\\
\bottomrule
\end{tabular*}
\label{tab:USVTrack-statistics}
\end{table}

Statistics for camera and radar modalities are summarized in TABLE \ref{tab:USVTrack-statistics}. The camera image modality contains 68,822 frames, with a total of 85,229 objects and 98 track IDs. Variations in capturing frequency and field of view between the camera and radar sensors result in differing statistical data. The radar point cloud modality includes 45,091 frames, with 44,271 targets and 89 track IDs. Furthermore, we conducted a comprehensive analysis of the radar point clouds by calculating the average values of attributes for each specific class. The number of average points is highly correlated with object size. Specifically, ships and vessels, being large objects, have a higher number of points, while boats, being smaller, have fewer points. Average power reflects the material properties of the objects. The three types of targets in USVTrack exhibit similar but distinct reflection intensities due to differences in the amount and proportion of metal and cement in their construction. Average velocity information is also instrumental in distinguishing between different types of objects. For example, ships and boats typically travel at slower velocities, while vessels exhibit relatively higher velocities.

\begin{figure}[h]
\centering
\subfloat[Size distribution]{
\includegraphics[width=0.48\columnwidth]{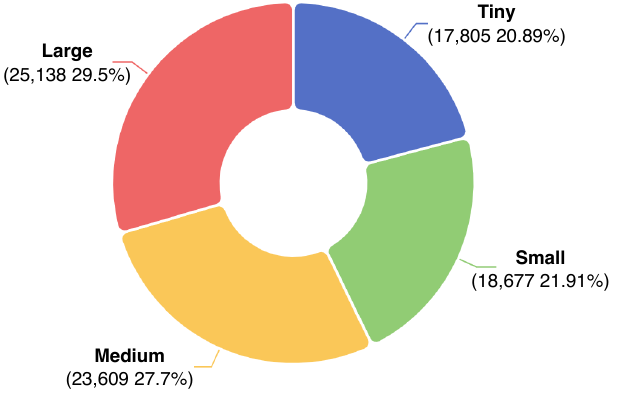}
\label{fig:size}
}
\quad
\hspace{-8mm}
\subfloat[Distance distribution]{
\includegraphics[width=0.48\columnwidth]{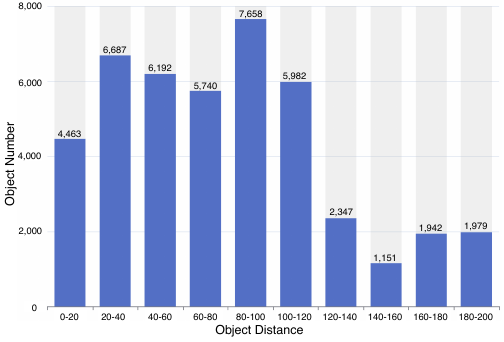}
\label{fig:distance}
}
\caption{Statistics of objects in USVTrack. (a) Wide range of object size. (b) Wide distribution of object distance.}
\label{fig:statistics}
\end{figure}

Compared to the MOT20 \cite{dendorfer2020mot20}, DanceTrack \cite{sun2022dancetrack} and VisDrone \cite{zhu2021detection} datasets, the object sizes in USVTrack vary due to the distance and the width of the waterways. We categorized the sizes of the object into four categories: those with a bounding box area less than 64 $\times$ 64 pixels are considered tiny, objects with an area between 64$^2$ and 128$^2$ pixels are referred to as small, those between 128$^2$ and 256$^2$ pixels are classified as medium, and areas greater than 256 $\times$ 256 pixels are deemed large. The number and distribution ratio of these four sizes of objects is shown in Fig. \subref*{fig:size}.
Leveraging the unique ranging capability of our radar sensor, we also analyzed the relationship between the number of objects at every 20 meters, as shown in Fig. \subref*{fig:distance}. Notably, the 80-100 meter interval contains a significantly higher number of objects compared to adjacent intervals, peaking at approximately 7,658 objects. Additionally, the two more distant intervals of 160-180 meters and 180-200 meters demonstrate the radar's ability to detect long-distance objects, with each interval containing over 1,900 objects.

\section{Benchmarks}
\label{Benchmarks}

In this section, our USVTrack dataset serves as resource for evaluating the performance of object detection and object tracking. By constructing the benchmarks using popular trackers and proposing a radar-camera matching method, we highlight the value, challenges and potential research directions posed by USVTrack for further research.

\subsection{Radar-Camera Matching (RCM)}
A significant challenge in waterborne object tracking lies in the frequent occurrence of identity switches caused by factors such as occlusions, motion disturbances, and heading angle variations. Inspired by ByteTrack's two-stage matching paradigm, which leverages both high- and low-score detection boxes to enhance trajectory coherence, we propose Radar-Camera Matching (RCM), a novel cross-modal association strategy designed to replace conventional IoU-based secondary matching in tracking frameworks. RCM integrates low-confidence camera detection boxes with radar detections to match remaining tracks, improving tracking robustness in complex scenarios.

When calculating the radar detections, as radar data suffers from noise and sparsity, we employ DBSCAN \cite{ester1996density} to cluster radar detection points, effectively merging multiple radar returns belonging to the same object into more reliable detection clusters. The process begins with clustering the raw radar detections using DBSCAN, which groups spatially proximate points while filtering out noise. From each resulting cluster, we extract dynamic attributes including average velocity and direction, which serve as robust motion features for subsequent tracking. These clustered radar detections are then fused with low-confidence camera detection bounding boxes by computing a matching cost based on spatial and dynamic consistency.

\begin{equation}
\begin{gathered}
C_{i, j}^{rcm}=\left\{\begin{array}{l}
1, \left(\hat{D}_{radar}>\theta_{rcm}\right) or \left(D_{iou}>\theta_{iou}\right)\\
\alpha\cdot \hat{D}_{radar} + (1 - \alpha) \cdot D_{iou}, \text{ otherwise }
\end{array}\right. \\
\end{gathered}
\label{eq:RCMatch}
\end{equation}
\begin{equation}
\hat{D}_{radar}=1-e^{-\lambda D_{radar }}
\label{eq:radar}
\end{equation}

As illustrated in Equation \ref{eq:RCMatch} and \ref{eq:radar}, RCM combines IoU geometric alignment and radar-based dynamic consistency, leveraging the complementary strengths of cameras (spatial precision) and radar (occlusion penetration). The radar-camera matching cost, denoted as $C_{i, j}^{rcm}$, quantifies the association likelihood between track $i$ and detection $j$.
$D_{iou}$ represents the IoU distance, measuring the spatial overlap between the predicted track bounding box and the detected bounding box. $\hat{D}_{radar}$ is the normalized Mahalanobis distance, which evaluates the consistency of radar-derived dynamic attributes (e.g., velocity and direction) between the track and detection. The parameter $\lambda$ controls sensitivity to radar distance, enabling fine-tuning of the radar's contribution to the matching process.

Thresholds $\theta_{iou}$ and $\theta_{rcm}$ are introduced to ensure robust association: $\theta_{iou}$ filters out geometrically inconsistent matches based on IoU, while $\theta_{rcm}$ discriminates between positive and negative radar-camera associations by evaluating the combined cost. The weighting factor $\alpha$ balances the contributions of geometric alignment and dynamic consistency, allowing the algorithm to adapt to varying environmental conditions and sensor reliability. This formulation not only enhances the robustness of cross-modal association but also addresses challenges such as occlusions, motion disturbances, and sparse radar data.

\subsection{Experimental Details}
All experiments are conducted on a high-performance computing setup, utilizing an NVIDIA RTX 4090 GPU paired with an Intel Xeon(R) Gold 6146 CPU @ 3.20GHz. The evaluation is performed on the test set of the USVTrack dataset to ensure rigorous assessment of the proposed methods.
 
\subsubsection{Object Detection}
To ensure a comprehensive evaluation, we selected representative object detectors spanning both two-stage and one-stage paradigms, including Faster R-CNN, and YOLO series from v8 to v12. Given the computational constraints and the need for real-time performance on USVs, we specifically chose lightweight versions of each detector. During the training phase, all images in USVTrack are resized as 640 $\times$ 640 pixels. All detectors are trained using default hyper-parameters for 100 epochs with early stop.

\subsubsection{Object Tracking}
Based on preliminary experiments, YOLOv9-N demonstrated the best performance and was selected as the default detector for tracking experiments. The confidence thresholds are set as follows: high-confidence detections are filtered at 0.5, while low-confidence detections are retained at 0.1 to ensure robustness in challenging scenarios. New tracks are initialized for detections with a confidence score higher than 0.2 if they do not match any existing tracks. To handle unmatched tracks, a buffer of 100 frames is employed, allowing for temporary retention and subsequent removal of tracks that fail to associate with new detections.
 
\subsubsection{RCM}
To validate the effectiveness of our proposed RCM module, we retained the original configurations of mainstream trackers and only replaced their second association phases with RCM.
The radar inconsistency threshold $\theta_{rcm}$ is set to 0.5, rejecting matches where radar measurements deviate from predictions. Similarly, the geometric misalignment threshold $\theta_{iou}$ is set to 0.5 to ensure spatial consistency between tracks and detections. Based on our extensive experience and experimental results, $\alpha$ is set to 0.7, prioritizing spatial precision while still leveraging radar-derived motion features. The parameter $\lambda$, which controls the sensitivity to radar distance, is empirically set to 1.2 to ensure optimal performance across diverse scenarios.

\subsection{Experimental Results}

\subsubsection{Object Detection}

\begin{table}[h]
\caption{\textbf{Object detection} results on USVTrack. In the Modalities column, C and R denote the modality from camera sensor and 4D radar sensor, respectively. Bold values indicate the optimal results, while underlined values denote the secondary outcomes.}
\center
\footnotesize
\begin{tabular*}{\linewidth}{
p{2.7cm}<{}
p{1.1cm}<{\centering}|
p{0.5cm}<{\centering}
p{0.55cm}<{\centering}
p{0.55cm}<{\centering}|
p{0.6cm}<{\centering}
}
\toprule
\bf{Detector} & \bf{Modalities} & \textbf{Ship} & \textbf{Boat} & \textbf{Vessel} & \bf{mAP}$\uparrow$ \\\midrule
Faster R-CNN \cite{ren2015faster} & C  & 85.5 & 66.2 & 71.3 & 73.9 \\
CenterNet \cite{zhou2019objects} & C & 91.8 & 83.2 & 84.4 & 85.7 \\
Deformable DETR \cite{zhu2020deformable} & C & 86.3 & 79.6 & 87.1 & 84.6 \\
YOLOX-T \cite{ge2021yolox} & C & 89.9 & 78.2 & 71.6 & 79.9 \\
YOLOv8-N \cite{yolov8} & C & 87.9 & 83.2 & 83.9 & 85.0 \\
YOLOv9-T \cite{wang2024yolov9} & C & \textbf{95.2} & \textbf{87.6} & \textbf{93.3} & \textbf{92.0} \\
YOLOv10-N \cite{wang2025yolov10} & C & 92.3 & 82.7 & 84.6 & 86.5 \\
YOLOv11-N \cite{yolov8} & C & 92.1 & \underline{85.2} & \underline{89.6} & \underline{89.0}\\
YOLOv12-N \cite{tian2025yolov12} & C & \underline{92.8} & 84.9 & 87.5 & 88.4 \\
\midrule
SAF-FCOS \cite{chang2020spatial} & C + R & 90.2 & 82.6 & 83.7 & 84.8 \\
CRF-Net \cite{nobis2019deep} & C + R & 89.7 & 78.6 & 72.8 & 86.4\\
RCDetect \cite{yao2024waterscenes} & C + R & 90.3 & 84.6 & 86.5 & 87.1 \\
\bottomrule
\end{tabular*}
\label{tab:usvtrack-detection-results}
\end{table}

The experimental results of the object detection are shown in TABLE \ref{tab:usvtrack-detection-results}. 
YOLOv9-N demonstrates state-of-the-art detection performance across all evaluation metrics, achieving the highest mAP$_{50\text{-}95}$ and mAP$_{50}$ scores among all evaluated detectors. Radar-camera fusion approaches exhibit competitive results, outperforming baseline detectors such as YOLOX-T and YOLOv8-N. However, when compared to current cutting-edge single-stage detectors, there remains a noticeable performance gap in both overall metrics and category-specific detection accuracy. 

\subsubsection{Object Tracking}

\begin{table*}[!h]
\caption{\textbf{Object tracking} results on USVTrack. In the Modalities column, C and R denote the modality from camera sensor and 4D radar sensor, respectively. Results for ``Ship," ``Boat," and ``Vessel" are derived from HOTA. Bold values indicate the optimal results, while underlined values denote the secondary outcomes.}
\center
\footnotesize
\begin{tabular*}{\linewidth}{
p{2.7cm}<{}
p{1.2cm}<{\centering}|
p{1.1cm}<{\centering}
p{1.1cm}<{\centering}
p{1.1cm}<{\centering}|
p{1.8cm}
p{1.8cm}
p{1.8cm}
p{1.35cm}}
\toprule
\bf{Tracker} & \bf{Modalities} & \textbf{Ship} & \textbf{Boat} & \textbf{Vessel} & \bf{HOTA}$\uparrow$ & \bf{MOTA}$\uparrow$ & \bf{IDF1}$\uparrow$ & \bf{IDSW}$\downarrow$ \\\midrule
StrongSORT \cite{du2023strongsort} & C & 19.196 & 10.535 & 19.889 & 16.478  & -635.8 & 10.374 & 1249 \\
Tracktor++ \cite{bergmann2019tracking} & C & 24.917 & 10.726 & 23.385 & 22.394 & -250.7 & 24.835 & 1082\\
FairMOT \cite{zhang2021fairmot} & C & 41.628 & 30.517 & 39.253 & 38.517 & 48.362 & 39.624 & 734\\
MOTRv2 \cite{zhang2023motrv2} & C & 45.209 & 32.715 & 48.615 & 44.305 & 58.317 & 46.815 & 413\\
MOTIP \cite{gao2024multiple} & C & 47.342 & 34.874 & 45.619 & 44.219 & 52.874 & 47.385 & 327 \\
Deep OC-SORT \cite{maggiolino2023deep} & C & \underline{54.652} & \underline{36.359} & \underline{54.370} & \underline{48.196} & 61.696 & \underline{53.887} & 287 \\
\midrule
BoT-SORT \cite{aharon2022bot} & C & 39.256 & 29.917 & 41.696 & 36.953  & 60.275 & 37.786 & 1106 \\
\rowcolor{lightgray} \qquad \qquad \qquad + \textbf{RCM} & C + R & 41.967 & 31.313 & 43.679 & 38.829 \scriptsize{(+1.876)} & 61.419 \scriptsize{(+1.144)} & 40.839 \scriptsize{(+3.053)} & 993 \scriptsize{(-113)} \\\midrule

ByteTrack \cite{zhang2022bytetrack} & C & 46.874 & 33.835 & 48.570 & 43.344 & 54.948 & 50.385 & 327 \\
\rowcolor{lightgray} \qquad \qquad \qquad + \textbf{RCM} & C + R & 50.498 & 34.542 & 52.399 & 44.298 \scriptsize{(+0.954)} & 55.284 \scriptsize{(+0.336)} & 52.123 \scriptsize{(+1.738)} & 295 \scriptsize{(-32)} \\\midrule

OC-SORT \cite{cao2023observation} & C & 52.122 & 35.669 & 52.625 & 47.348  & 62.557 & 51.790 & 415 \\
\rowcolor{lightgray} \qquad \qquad \qquad + \textbf{RCM} & C + R & 53.377 & \textbf{37.523} & \textbf{54.940} & 48.098 \scriptsize{(+0.750)} & 62.986 \scriptsize{(+0.429)} & 53.498 \scriptsize{(+1.708)} & 373 \scriptsize{(-42)} \\\midrule

Hybrid-SORT \cite{yang2024hybrid} & C & 52.571 & 35.382 & 51.318 & 47.208 & \underline{63.525} & 52.951 & \underline{207} \\
\rowcolor{lightgray} \qquad \qquad \qquad + \textbf{RCM} & C + R & \textbf{54.879} & 36.213 & 53.476 & \textbf{48.302} \scriptsize{(+1.094)}  & \textbf{63.739} \scriptsize{(+0.214)} & \textbf{54.199} \scriptsize{(+1.248)} & \textbf{186}  \scriptsize{(-21)} \\
\bottomrule
\end{tabular*}
\label{tab:USVTrack-tracking-results}
\end{table*}

Comparative results of popular JDT and TBD paradigms is summarized in TABLE \ref{tab:USVTrack-tracking-results}. 
Without the second-stage IoU matching, Deep OC-SORT achieves the highest HOTA score of 48.196 and IDF1 score of 53.887. 
The MOTA metric for StrongSORT and Trackor++ are negative, indicating that the number of false positives, false negatives, and ID switches far exceeds the number of ground truth.

For trackers incorporating the second-stage IoU matching, we replaced the traditional IoU matching with our proposed RCM method. Experimental results show that the RCM improves the performance of four two-stage associated trackers, including metrics such as HOTA, MOTA, IDF1 and IDSW. For instance, in the Hybrid-SORT method, the introduction of RCM led to an increase of 1.094 (from 47.208 to 48.302) in the HOTA metric, and a reduction of 21 (from 207 to 186) in IDSW. These improvements validate the effectiveness of RCM in enhancing object association and reducing mismatches.

\begin{figure*}[!h]
\centering
\includegraphics[width=1\linewidth]{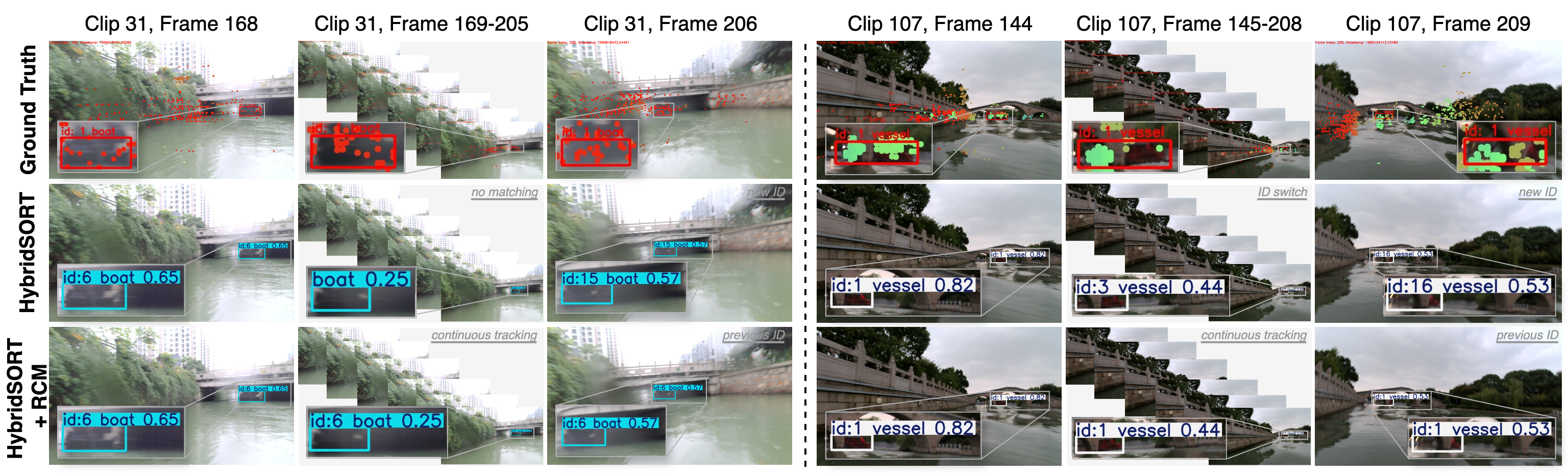}
\caption{Visualization of object tracking on USVTrack. The first row shows the ground truth, the second displays  results from HybridSORT, and the third row presents result from HybridSORT with our RCM module. With RCM integrated, trackers exhibit reduced IDSW and maintain continuous tracking.}
\label{fig:RCTrack-visualization}
\end{figure*}

\subsubsection{Visualization Analysis}
To visually evaluate the effectiveness of the RCM module, we conducted visualization analysis in two typical scenarios based on the HybridSORT framework. As depicted in Fig. \ref{fig:RCTrack-visualization}, the first row displays the visual image, radar point cloud, and annotated ground truth information, including object categories and tracking IDs. The second row presents the tracking results from the baseline HybridSORT, revealing issues such as matching failures and ID switches, particularly in scenarios involving camera shake and distant small targets. In contrast, the integration of our RCM module, shown in the third row, demonstrates consistent tracking performance, effectively maintaining ID continuity even under challenging conditions. These results highlight RCM's ability to enhance the reliability and stability of object tracking in complex environments.

\section{Discussion}

\subsection{Radar-Camera Tracking in Inland Waterways}
Radar-camera fusion methods for waterway applications are still in their early stages. While our proposed RCM method provides a promising direction, there is ample room for innovation in radar-camera fusion techniques. Developers could explore joint radar-camera tracking strategies to further advance the field. Enhancing radar-camera object detection could also improve the stability and accuracy of subsequent tracking tasks. Specific challenges, such as ID switches caused by viewpoint changes and the tracking of small targets, remain critical issues. Leveraging advanced feature representations or incorporating temporal consistency models could help mitigate these problems and improve tracking robustness.

\subsection{Limitations of USVTrack for Autonomous Driving}
Our USVTrack dataset, while comprehensive in certain aspects, still has some limitations. Firstly, it primarily focuses on inland waterways, and annotations are limited to major target categories such as boat, ship, and vessel. Other categories, such as kayak and ferry, are present in USVTrack but were not annotated due to their low sample size, which may restrict the dataset's applicability to scenarios involving a broader range of watercraft. Additionally, the distribution of targets in inland waterway scenes is uneven, with some scenes containing only a single boat on a wide lake, which fails to adequately represent dense object scenarios.

\section{Conclusion}
With the concept of new generation waterborne transportation systems, navigation intelligence has emerged as a critical research area. In this work, we present the first 4D radar-camera fusion dataset specifically designed for USVs in waterborne environments. Our dataset provides a wealth of real-world data, encompassing diverse waterway scenarios, varying weather conditions, and dynamic lighting, to meet the high-quality data requirements of autonomous driving in inland waterways.
In addition to the dataset, we propose a simple yet effective radar-camera matching method named RCM, which can be seamlessly integrated into trackers with two-stage associations. The RCM approach effectively reduces the frequency of identity switches, enhancing the stability and reliability of object tracking. By improving the association between radar and camera data, our method addresses one of the key challenges in multi-modal fusion for waterborne applications.
Overall, our work lays a solid foundation for future research in waterborne intelligence and multi-modal fusion. By leveraging our dataset and method, researchers can further advance the capabilities of USVs and contribute to the realization of new generation waterborne transportation systems.

\bibliographystyle{IEEEtran}
\bibliography{citations,others}

\end{document}